\documentclass[letterpaper, 10 pt, conference]{IEEEtran}

\PassOptionsToPackage{hyphens}{url}\usepackage{hyperref}
\usepackage{blindtext}
\usepackage{eso-pic}
\IEEEoverridecommandlockouts
\usepackage{cite}
\usepackage{amsmath,amssymb,amsfonts}
\usepackage{algorithmic}
\usepackage{graphicx}
\usepackage{textcomp}
\usepackage{xcolor}

\usepackage[export]{adjustbox}
\usepackage[utf8]{inputenc} % allow utf-8 input
\usepackage[T1]{fontenc}    % use 8-bit T1 fonts
\usepackage{hyperref}       % hyperlinks
\usepackage{booktabs}       % professional-quality tables

\usepackage{makecell}
\usepackage{amsfonts}       % blackboard math symbols
\usepackage{nicefrac}       % compact symbols for 1/2, etc.
\usepackage{microtype}      % microtypography
\usepackage{xcolor}         % colors
\usepackage{amsmath,amssymb,multicol,latexsym}
\usepackage{pgfplots,pgfplotstable}
\pgfplotsset{compat=1.16}
\usepackage{acro}
\usepackage{tikz}
\usepackage{comment}
\usetikzlibrary{shapes,
 	automata,
 	arrows.meta,arrows,
 	chains,
 	matrix,
 	backgrounds,
 	fit,
 	patterns,
 	decorations.markings,
 	svg.path,
 	shapes.multipart,
 	shapes.symbols,
 	external,
 	shadows,
 	positioning,
 	calc,
 	decorations,
 	scopes,
 	patterns,
 	patterns.meta,
 	bending}
 \usetikzlibrary{external}

\usepackage{hyperref}
\usepackage{cleveref}
\usepackage{flushend}

\DeclareMathAlphabet{\mathcal}{OMS}{cmsy}{m}{n}

\usepackage{url}
\usepackage{graphicx}
\usepackage{svg}
\usepackage{numprint,fullpage}
\usepackage{booktabs}
\usepackage{multirow}

\usepackage{adjustbox}

\Crefname{equation}{Eq.}{Eqs.}
\Crefname{figure}{Fig.}{Figs.}
\Crefname{tabular}{Tab.}{Tabs.}
\usepackage{tabu}
\usepackage{anysize} % Soporte para el comando \marginsize
\graphicspath{{figures/}}
\def\BibTeX{{\rm B\kern-.05em{\sc i\kern-.025em b}\kern-.08em
    T\kern-.1667em\lower.7ex\hbox{E}\kern-.125emX}}
\usepackage{url}

\begin{document}
\title{Clustering-based Criticality Analysis for Testing of Automated Driving Systems}

\author{\IEEEauthorblockN{Barbara~Schütt, Stefan~Otten, Eric~Sax}
\IEEEauthorblockA{
FZI Research Center for Information Technology \\
Karlsruhe, Germany
Email: \{schuett, otten, sax\}@fzi.de}}

\maketitle
\thispagestyle{empty}
\pagestyle{empty}

%%%%%%%%%%%%%%%%%%%%%%%%%%%%%%%%%%%%%%%%%%%%%%%%%%%%%%%%%%%%%%%%%%%%%%%%%%%%%%%%
\begin{abstract}

With the implementation of the new EU regulation 2022/1426 regarding the type-approval of the automated driving system (ADS) of fully automated vehicles, scenario-based testing has gained significant importance in evaluating the performance and safety of advanced driver assistance systems and automated driving systems.
However, the exploration and generation of concrete scenarios from a single logical scenario can often lead to a number of similar or redundant scenarios, which may not contribute to the testing goals.

This paper focuses on the the goal to reduce the scenario set by clustering concrete scenarios from a single logical scenario.
By employing clustering techniques, redundant and uninteresting scenarios can be identified and eliminated, resulting in a representative scenario set.
This reduction allows for a more focused and efficient testing process, enabling the allocation of resources to the most relevant and critical scenarios.
Furthermore, the identified clusters can provide valuable insights into the scenario space, revealing patterns and potential problems with the system's behavior.

\end{abstract}

%%%%%%%%%%%%%%%%%%%%%%%%%%%%%%%%%%%%%%%%%%%%%%%%%%%%%%%%%%%%%%%%%%%%%%%%%%%%%%%%
\section{Introduction}
%The development of advanced driver assistance systems (ADAS) and automated driving systems (ADS) has been a significant focus of the automotive industry in recent years. 
%During the last years, the consulting company Gartner recognized the emergence of \textit{autonomous things}, \textit{hyperautomation}, or \textit{autonomic systems} as a hot topic \cite{Gartner, Gartner2}.
%These topics include systems that not only make autonomous decisions but also have the ability to adapt and change their behavior according to their environment. 
%One type of autonomous or autonomic system is automated vehicles. 
%However, a major challenge besides their development is to ensure that the system is safe enough to be approved and permitted on public roads. 
%In system and software development, testing is an important step to develop reliable systems.
%In particular, i
In the automotive industry, software validation and verification are an integral part of the development process to make sure requirements are met and the system fulfills the intended use cases \cite[p.~76]{wood_safety_2019}. 
According to \cite{wachenfeld2016release}, real world test drives, known as distance-based testing, are a valid testing method and give the most reliable results. 
However, distance-based testing is not feasible: 
6 billion kilometers are necessary to make a statistically significant statement about a system's safety.

One widely discussed alternative testing approach is scenario-based testing, which involves taking field test drives into simulation environments. 
There predefined scenarios serve as a basis for the derivation of relevant test cases in automated assessment, thus reducing the number of real test drives needed. 
The proper representation of scenarios during the development process supports a seamless development and testing of automated driving functions, as well as the specification of requirements. 

Additionally, the importance of scenario-based testing has been further increased in 2022. 
The EU implementation regulation \cite{EUtypeapprovannex} 2022/1426 for type-approval of the Automated Driving System (ADS) came into effect on August 5, 2022, and is now mandatory to be followed in all EU member states. 
%It lists a number of methods for the overall compliance assessment of an automated driving system. 
The implementation regulation states that a minimum set of traffic scenarios shall be used in case these scenarios are relevant to the automated driving system's (ADS) operational design domain.
For some functions, e.g., lane keep assistance, it even defines the minimum set of test scenarios. 
Further, specification of scenarios can include parameter ranges, where only a subset of these ranges may provide insight into the performance of an automated driving function or hold critical scenarios, and introducing new parameters or parameter ranges in a scenario increases the number of scenarios exponentially.

The main contributions of this work encompasses three topics of interest:
\begin{itemize}
    \item[(1)] Evaluation of Criticality: One focus lies on the evaluation of criticality within the scenario space of one logical scenario, identifying critical areas.
    \item[(2)] Two Scenario Clustering Approaches: The first approach is behavior-based clustering, which categorizes scenarios based on the similarity of actors' behaviors, to identify clusters with similar outcomes. The second approach is criticality-based clustering, where scenarios are clustered based on their course of criticality, facilitating the grouping of high-risk and low-risk scenarios.
    \item[(3)] Scenario Set Reduction: By employing the aforementioned clustering approaches, redundant and uninteresting scenarios are eliminated, resulting in a more concise and manageable scenario set. 
    \item[4)] Reduction of computation time compared to a grid-based approach for logical scenarios.
\end{itemize}

In \Cref{sec:related_work}, terms and concepts related to scenario-based testing are presented.
In \Cref{sec:experiments}, we propose scenario exploration with a novel analysis concept, which forms the basis of our experiments.
The conducted experiments are then summarized and explained in detail. 
Finally, we conclude and discuss possible future research directions in \Cref{sec:conclusion}.
\section{Related work}
\label{sec:related_work}
\begin{figure}[tbh]
    \centering
    \begin{adjustbox}{center,width=0.4\textwidth}
		\begin{tikzpicture}[sideArrow/.style={signal, font=\small, fill=#1!20, signal pointer angle=100, align=center},]

% Levels of Abstraction
    \tikzstyle{box}=[draw,minimum height=17pt, minimum width=238pt, semithick, inner sep=5pt, outer sep=0pt, anchor=north]
	\node[box, align=center, anchor=center] at (0,0) (Label2){Abstraction Levels};
	
	\tikzstyle{box}=[draw,minimum height=15pt, minimum width=78pt, semithick, inner sep=0pt, outer sep=0pt, fill=black!12, anchor=north]
	\node[box, align=center, anchor=north] at (0.0, -0.35) (LevelLog){Logical};

	\tikzstyle{box}=[draw, minimum height=15pt, minimum width=78pt, semithick, inner sep=0pt, outer sep=0pt, fill=black!12, anchor=north]
	\node[box, align=center, anchor=east] at ($(LevelLog.west) + (down:0pt) + (left:2pt)$) (LevelFunc){Functional};
	
	\tikzstyle{box}=[draw, minimum height=15pt, minimum width=78pt, semithick, inner sep=0pt, outer sep=0pt, fill=black!12, anchor=north]
	\node[box, align=center, anchor=west] at ($(LevelLog.east) + (down:0pt) + (right:2pt)$) (LevelConc){Concrete};

	\tikzstyle{box}=[draw,minimum height=40pt, minimum width=78pt, semithick, inner sep=2pt, outer sep=0pt,execute at begin node=\setlength{\baselineskip}{10pt},anchor=north]
	\node[box, align=center, anchor=north] at ($(LevelFunc.south) + (down:2pt) + (right:0pt)$) (LevelFuncDesc){\footnotesize Scenario description  \\ \footnotesize via linguistic notation \\ \footnotesize e.g., natural language.};

	\tikzstyle{box}=[draw,minimum height=40pt, minimum width=78pt, semithick, inner sep=2pt, outer sep=0pt,execute at begin node=\setlength{\baselineskip}{10pt},anchor=north]
	\node[box, align=center, anchor=north] at ($(LevelLog.south) + (down:2pt) + (right:0pt)$) (LevelLogDesc){\footnotesize Scenario description  \\ \footnotesize with help of parameter \\ \footnotesize ranges or probability \\ \footnotesize distributions.};
	
	\tikzstyle{box}=[draw,minimum height=40pt, minimum width=78pt, semithick, inner sep=2pt, outer sep=0pt,execute at begin node=\setlength{\baselineskip}{10pt},anchor=north]
	\node[box, align=center, anchor=north] at ($(LevelConc.south) + (down:2pt) + (right:0pt)$) (LevelConcDesc){\footnotesize Scenario description  \\ \footnotesize with given concrete \\ \footnotesize parameter for each \\ \footnotesize variable};
	
	\tikzstyle{box}=[draw,minimum height=45pt, minimum width=78pt, semithick, inner sep=2pt, outer sep=0pt,execute at begin node=\setlength{\baselineskip}{10pt},anchor=north]
	\node[box, align=center, anchor=north] at ($(LevelFuncDesc.south) + (down:2pt) + (right:0pt)$) (LevelFuncExp){\footnotesize Setting: intersection  \\ \footnotesize Ego vehicle takes \\  \footnotesize right turn. \\ \footnotesize Bike crosses street.};

	\tikzstyle{box}=[draw,minimum height=45pt, minimum width=78pt, semithick, inner sep=2pt, outer sep=0pt,execute at begin node=\setlength{\baselineskip}{10pt},anchor=north]
	\node[box, align=center, anchor=north] at ($(LevelLogDesc.south) + (down:2pt) + (right:0pt)$) (LevelLogExp){\footnotesize Lane width: \\\scriptsize $[2.5-3.5m]$  \\ \footnotesize Ego \scriptsize $v \in [2.0-3.5\frac{m}{s}]$  \\ \footnotesize Bike \scriptsize $v \in [1.0-2.5\frac{m}{s}]$ };
	
	\tikzstyle{box}=[draw,minimum height=45pt, minimum width=78pt, semithick, inner sep=2pt, outer sep=0pt,execute at begin node=\setlength{\baselineskip}{10pt},anchor=north]
	\node[box, align=center, anchor=north] at ($(LevelConcDesc.south) + (down:2pt) + (right:0pt)$) (LevelConcExp){\footnotesize  Lane width:\scriptsize $3.0m$  \\ \footnotesize Ego \scriptsize $v=2.0\frac{m}{s}$ \\ \footnotesize Bike \scriptsize $v=1.0\frac{m}{s}$};

    \coordinate [] (A) at (-4.2, -4.7);
  \coordinate [] (B) at (-4.2, -4.1);
  \coordinate [] (C) at (4.2, -4.1);
  \node[align=center, anchor=north] at (-2.7, -4.1) (lab1){Level of abstraction};
  
  \coordinate [] (D) at (4.2, -4.2);
  \coordinate [] (E) at (4.2, -4.8);
  \coordinate [] (F) at (-4.2, -4.8);
  \node[align=center, anchor=north] at (2.6, -4.35) (lab1){Number of scenarios};

  \draw [] (A) -- (C) -- (B) -- (A);
  \draw [] (D) -- (E) -- (F) -- (D);
	
\end{tikzpicture}
	\end{adjustbox}
    \caption{All 3 levels of scenario abstraction with examples \cite{menzel2018scenarios}.}
    \label{fig:abstraction}
\end{figure}

In this work, the term \textit{scenario} is used as defined by Steimle et al. \cite{steimle2021toward}. 
A scenario is a sequence of scenes that describes the temporal development of the behavior of different actors within this sequence, where a scene is a snapshot of a traffic constellation.

\subsection{Scenario Abstraction}
According to Menzel \textit{et al.} \cite{menzel2018scenarios}, scenario representation has three levels of abstraction that are shown in \Cref{fig:abstraction}.
The first level is the most abstract one, the \textit{functional} level, which uses natural language to describe scenarios and focuses on creating easily understandable scenarios for discussion. 
The second level is the \textit{logical} level and refines the functional level by using state-space variables to describe the scenario in more detail, including parameters such as road width, vehicle positions, or weather conditions. 
The most detailed level is the \textit{concrete} level, which specifies each parameter with a concrete value, potentially generating infinite scenarios from a single logical scenario depending on the size and step size of the parameter ranges or distribution.

In theory, a logical scenario can yield infinite concrete scenarios that cannot be exhaustively simulated.
A simple and obvious strategy would be to use a grid for all parameter ranges.
However, this method has been criticized for its computational effort and shown to overlook scenarios that are more critical than those simulated, as demonstrated by Mori  \textit{et al.} \cite{mori2022inadequacy}.

\subsection{Scenario Exploration and Acquisition}
In general, the process of finding suitable scenarios is a crucial step in defining new test cases to assess the safety of ADS. 
Ideally, these scenarios provide realistic critical and relevant situations for testing the ADS' performance.
%This indicates, that not only the ADS' performance has to be evaluated, but also the scenario quality and criticality \cite{schutt2022taxonomy}.
Typical criticality metrics are time-to-collision (TTC) \cite{hayward_near_1972}, post-encroachment time (PET) \cite{Allen.1978}, worst time-to-collision (WTTC) \cite{wachenfeld_worst-time--collision_2016}, Euclidean distance, or traffic quality (TQ) \cite{schutt2023inverse}.

Scenarios can be found in various ways: 
Adapting scenario parameters, extracting scenarios from naturalistic driving data and other sources, procedural generation, or human experts \cite{schuett20231001ways}.
Langner \textit{et al.} \cite{langner2019logical} present an approach to extract dynamic-length-segments containing a single scenario, enrich them with a feature vector, cluster them, and create a logical scenario catalog representing corner cases and common scenarios with an accumulated total length for each scenario.
Another category is scenario exploration, the search for critical and relevant concrete scenarios within a logical scenario.
\cite{bussler2020application} \textit{et al.} employ evolutionary learning techniques to discover relevant parameter sets that yield significant scenarios within logical scenarios.
They utilize Euclidean distance and TTC as fitness evaluation to identify more critical scenarios.
\cite{baumann2021automatic} \textit{et al.} propose an alternative approach using reinforcement learning combined with the metrics headway and time-to-collision to generate new test cases.
Additionally, \cite{abeysirigoonawardena2019generating} \textit{et al.} use Bayes Optimization and Euclidean distance to generate training scenarios for a driving function aimed at learning to avoid hitting pedestrians through reinforcement learning. 
However, their methodology does not yield a scenario set suitable for testing since their approach always relies on current state of the driving function which evolves over the course of the experiments.
Further approaches to discover new scenarios include extracting scenarios from recorded data sets as demonstrated by \cite{king2021capturing} \textit{et al.} and \cite{zofka2015data} \textit{et al.} or by experts planning and designing scenarios from scratch.
The scenario exploration and selection of simulated scenarios is based on the authors' previous work, where Bayes optimization with Gaussian processes is used to identify critical scenarios \cite{schutt2022application}.

\subsection{Scenario Clustering}
 %clustering is a widely used technique to group similar scenarios.
Scenario clustering is typically used to identify concrete scenarios within a logical scenario, where similar sequences of scenes, trajectories, or concrete scenarios are summarized into one scenario description with a range of parameters \cite{nalic2020scenario}.
A common approach for grouping is based on vehicle trajectories that can be extracted from various types of data, e.g., recorded naturalistic driving data sets  \cite{langner2019logical}.

\subsection{Simulation Tools}
A variety of commercial and open-source tools are available for automotive simulation, offering diverse functionalities and capabilities.
%Examples for commercial tools are \cite{dSPACE} and IPG \cite{carmaker}.
Both simulation tools provide modules for map and scenario creation, sensor models and dynamic models, to name some examples. 
A further tool is Carla, an open-source simulator that features a growing community and is based on the Unreal game engine \cite{Dosovitskiy17}.
It offers several additional modules, including a scenario tool with its own scenario format, a graphical tool for creating scenarios, a ROS/ROS2-bridge, and SUMO support.
SUMO is an open-source software tool for modeling microscopic traffic simulation \cite{SUMO2018}.
It enables the modeling of large-scale traffic scenarios, facilitating evaluations of traffic light cycles, emissions (such as noise and pollutants), traffic forecasting, and more.

%\subsection{Scenario Clustering and Analysis}
%After the optimization process, a we conducted a comprehensive analysis of the simulation results to obtain a better understanding %of the logical scenario outcomes. 
%This analysis was completed in several steps:
%\begin{itemize}
%    \item Computation of the distance matrix for all scenarios using dynamic time warping \cite{bellman1959adaptive} as a distance measure for vehicle trajectories or course of the criticality metrics during a scenario.
%    \item Dimensionality reduction using Kernel PCA \cite{mika1998kernel}.
%    \item Clustering of scenarios using DBSCAN \cite{ester1996density}.
%\end{itemize}
\section{Scenario Exploration and Analysis}
\label{sec:experiments}

Scenario exploration is an optimization process and aims to identify all critical concrete scenarios involving the ego vehicle and other actors within a logical  scenario.
Further, the analysis of the simulation recordings and results from this logical scenario seeks to find a subset of relevant scenarios and information about the ADS's performance within this logical scenario.

\subsection{Optimization Workflow}
%\begin{figure}[t!]
%    \centering
%       \includegraphics[width=0.52\linewidth]{img/scenario_exploartion_activity2.pdf}
%\caption{Activity diagram of the optimization workflow.}
%\label{fig:activity_uml}
%\vspace{0ex}
%\end{figure}
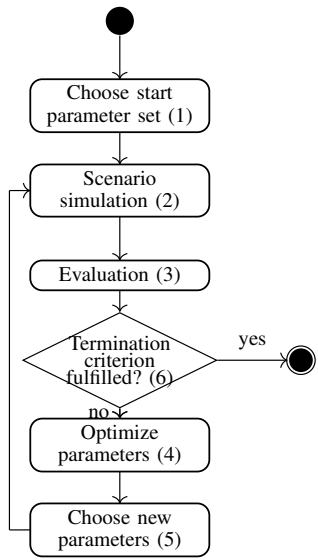
\begin{figure}[tbh]
    \centering
    \begin{adjustbox}{center,width=0.55\textwidth}
		% \ u s e t i k z l i b r a r y { shapes }
\begin{tikzpicture}
\tikzstyle{activity}=[draw,rounded corners, minimum height=10pt, minimum width=60pt, semithick, inner sep=2pt, outer sep=0pt, anchor=north];
\tikzstyle{decision}=[shape aspect=2,diamond,draw,minimum width=65pt,minimum height=32pt];
\tikzstyle{end}=[draw,double=white,circle,inner sep=1pt,minimum height=0.3cm,fill=black];
\tikzstyle{start}=[circle,minimum width=0.3cm,minimum height=0.3cm, fill=black];
\tikzstyle{textbox}=[minimum width=10pt, semithick, inner sep=2pt, outer sep=0pt, anchor=north];
\node[start,align=center] (start) {};
\node[activity, below of=start,align=center] (action1) {\scriptsize Choose start\\[-1ex]\scriptsize parameter set (1)};
\node[activity, below of=action1,align=center] (action2) {\scriptsize Scenario\\[-1ex]\scriptsize simulation (2)};
\node[activity, below of=action2,align=center] (action3) {\scriptsize Evaluation (3)};
\node[decision, below of=action3,align=left](decision1){};
\node[textbox, align=center] at ($(decision1.north) + (down:6pt) + (right:0pt)$)(decision1text){\scriptsize Termination \\ [-1.4ex]\scriptsize criterion \\ [-1.4ex]\scriptsize fulfilled? (6)};
%\node[textbox, align=center] at ($(decision1.north) + (down:8pt) + (right:0pt)$)(decision1text){};
\node[activity, below of=decision1,align=center] (action4) {\scriptsize Optimize\\[-1ex]\scriptsize parameters (4)};
\node[activity, below of=action4,align=center] (action5) {\scriptsize Choose new\\[-1ex]\scriptsize parameters (5)};
\node[end]at ($(decision1.east) + (down:0pt) + (right:28pt)$)(end){};
\draw [->](start) -- (action1);
\draw [->](action1) -- (action2);
\draw [->](action2) -- (action3);
\draw [->](action3) -- (decision1);
\draw [->](decision1) -- node [left]{\scriptsize  no} (action4);
\draw [->](decision1) -- node [above]{\scriptsize  yes} (end.west);
\draw [->](decision1) -- (action4);
\draw [->](action4) -- (action5);
\node (ctrl2) [anchor=center]at (-1.3,-2.0) {}; 
\node (ctrl1) [anchor=center]at (-1.3,-6.0) {}; 
\draw[to path={-- (\tikztotarget)}](action5.west) edge[-] (ctrl1.center);
\draw[to path={-- (\tikztotarget)}](ctrl1.center) edge[-] (ctrl2.center);
\draw[to path={-- (\tikztotarget)}](ctrl2.center) edge[->] (action2.west);
%\draw [->](decision1) -| (end.west);
\end{tikzpicture}
	\end{adjustbox}
    \caption{Activity diagram of the scenario exploration workflow.}
    \label{fig:activity_uml}
\end{figure}
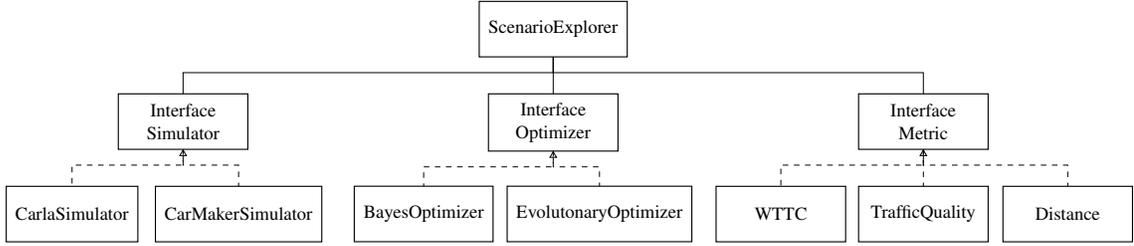
\begin{figure*}[bht!]
    \centering
    \begin{adjustbox}{center,width=0.7\textwidth}
		\begin{tikzpicture}[sideArrow/.style={signal, font=\small, fill=#1!20, signal pointer angle=100, align=center},]
\tikzstyle{box}=[draw, minimum height=30pt, minimum width=70pt, semithick, inner sep=5pt, outer sep=0pt, anchor=north];
\tikzstyle{textbox}=[minimum width=10pt, semithick, inner sep=2pt, outer sep=0pt, anchor=north, fill=white];

\node[box, align=center, draw=black!] at (0,0) (explorer){ScenarioExplorer};
\node[box, align=center, draw=black!] at ($(explorer.south) + (down:20pt) + (left:200pt)$) (simulator){Interface \\ Simulator};
\node[box, align=center, draw=black!] at ($(explorer.south) + (down:20pt) + (left:0pt)$) (optimizer){Interface \\ Optimizer};
\node[box, align=center, draw=black!] at ($(explorer.south) + (down:20pt) + (right:200pt)$) (metric){Interface \\ Metric};

\node[box, align=center, draw=black!] at ($(simulator.south) + (down:20pt) + (left:60pt)$) (carla){CarlaSimulator};
\node[box, align=center, draw=black!] at ($(simulator.south) + (down:20pt) + (right:30pt)$) (cm){CarMakerSimulator};
\node[box, align=center, draw=black!] at ($(optimizer.south) + (down:19pt) + (left:70pt)$) (bayes){BayesOptimizer};
\node[box, align=center, draw=black!] at ($(optimizer.south) + (down:19pt) + (right:25pt)$) (evol){EvolutonaryOptimizer};
\node[box, align=center, draw=black!] at ($(metric.south) + (down:20pt) + (left:77pt)$) (wttc){WTTC};
\node[box, align=center, draw=black!] at ($(metric.south) + (down:20pt) + (right:0pt)$) (tq){TrafficQuality};
\node[box, align=center, draw=black!] at ($(metric.south) + (down:20pt) + (right:78pt)$) (dist){Distance};

\draw[] (explorer.south) |-++(0,-0.3)-|(simulator.north);
\draw[] (explorer.south) |-++(0,-0.3)-|(optimizer.north);
\draw[] (explorer.south) |-++(0,-0.3)-|(metric.north);

\draw[dashed,{Triangle[open]}-] (simulator.south) |-++(0,-0.3)-|(carla.north);
\draw[dashed] (simulator.south) |-++(0,-0.3)-|(cm.north);
\draw[dashed,{Triangle[open]}-] (optimizer.south) |-++(0,-0.3)-|(bayes.north);
\draw[dashed,] (optimizer.south) |-++(0,-0.3)-|(evol.north);
\draw[dashed,{Triangle[open]}-] (metric.south) |-++(0,-0.3)-|(wttc.north);
\draw[dashed,] (metric.south) |-++(0,-0.3)-|(tq.north);
\draw[dashed,] (metric.south) |-++(0,-0.3)-|(dist.north);

\end{tikzpicture}
	\end{adjustbox}
    \caption{Class diagram of the simulation framework setup.}
    \label{fig:class_uml}
\end{figure*}

The optimization process is an iterative technique as depicted in Fig.~\ref{fig:activity_uml}. 
Initially, a start parameter set is chosen (1), and the simulation is performed (2). 
Subsequently, the results are evaluated (3), and an optimization algorithm (4) selects a new parameter set (5) to be simulated again (2). 
This procedure is repeated until a termination criterion is satisfied (6). 
A similar framework was already use by Schütt \textit{et al.} \cite{schutt2022application}.

%\subsection{Simulator Framework Setup}
%\begin{figure*}[t!]
%    \centering
%       \includegraphics[width=0.99\linewidth]{img/class_diagram.png}
%\caption{Class diagram of the simulation framework setup.}
%\label{fig:class_uml}
%\vspace{0ex}
%\end{figure*}

\Cref{fig:class_uml} shows the class diagram of the simulation framework setup.
The explorer needs a simulator, one or more criticality metrics, and an optimizer.
IPG CarMaker 8.0.2 \cite{carmaker} or the Carla simulator version 0.9.13 with the scenario runner \cite{Dosovitskiy17} were used for all following experiments.
All scenarios were detailed in CarMaker's own or, in the case of Carla, in the OpenSCENARIO format \cite{assam_openscenario}. 
The open-source project, Common Bayesian Optimization Library (COMBO), was employed to perform the optimization experiments \cite{ueno2016combo}. 
This library uses Bayesian optimization with automatic hyperparameter tuning, Thompson sampling to choose the next best candidate, and random feature maps to improve performance. 
The Bayesian optimization and Gaussian processes were utilized throughout this work, although other optimization algorithms could also be used. 
However, it is worth noting that this work's primary focus is not solely on the optimization itself.

\subsection{Optimization Setup}
\label{sec:opt_steup}

\begin{figure}[tbh]
    \centering
    \begin{adjustbox}{center,width=0.65\linewidth,margin*=-5ex 0ex 0ex 1ex}
		\tikzsetnextfilename{scenarios}
\begin{tikzpicture}[
street/.style={rectangle, minimum width=40pt, minimum height=160pt, fill=#1!30, draw=#1!30, text=black},
street2/.style={rectangle, minimum width=40pt, minimum height=80pt, fill=#1!30, draw=#1!30, text=black},
truck/.style={rectangle, minimum width=15pt, minimum height=50pt, fill=#1!30, draw=#1!70, text=black},
ego/.style={rectangle, minimum width=15pt, minimum height=25pt, text=black, fill=white, draw=black},
triangle/.style = {draw=#1!70, regular polygon, regular polygon sides=3, inner sep=0.1cm}
]

% scenario 1
\node[street=gray, anchor=west, align=center] (IS1) {};
\node[anchor=north, text width=95pt, align=center] at ($(IS1.north west) + (up:5pt) + (left:60pt)$) (ego1text){\Large (a)};
\node[street=gray, anchor=west, align=center, rotate=90] at  ($(IS1.center) + (down:20pt) + (left:0pt)$) (IS1b) {};

\node[circle,minimum size=12pt,inner sep=0pt] at ($(IS1.east) + (up:60pt) + (left:8pt)$) (cd1) {};
\node[circle,minimum size=12pt,inner sep=0pt] at ($(IS1b.south) + (down:10pt) + (left:20pt)$) (etd1) {};
\node[circle,minimum size=12pt,inner sep=0pt] at ($(IS1.east) + (down:37pt) + (right:8pt)$) (pd1) {};

\node[truck=blue, anchor=west, align=center] at  ($(IS1.west) + (up:50pt) + (right:4pt)$) (truck1) {T};
\node[triangle=blue, anchor=north, align=center, rotate=180] at  ($(truck1.south) + (up:0pt) + (left:0pt)$) (truck1b) {};
\node[ego, anchor=south, align=center] at  ($(IS1.south) + (up:30pt) + (right:10pt)$) (ego1) {};
\node[triangle=black, anchor=north, align=center] at  ($(ego1.north) + (up:0pt) + (left:0pt)$) (ego1b) {};
\node[anchor=north, text width=95pt, align=center] at ($(ego1b.south) + (down:0pt) + (right:0pt)$) (ego1text){E};
\node[ego, anchor=south, align=center, rotate=90] at  ($(IS1.east) + (up:10pt) + (right:40pt)$) (car1) {};
\node[triangle=black, anchor=north, align=center, rotate=90] at  ($(car1.north) + (up:0pt) + (left:0pt)$) (car1b) {};
\node[anchor=north, text width=95pt, align=center, ->] at ($(car1b.east) + (up:2pt) + (right:10pt)$) (car1text){C};
%edge[arrows = {-Stealth[scale=2]}, dashed, controls=+(right:5mm) and +(down:5mm)](cd1.south);
\node[draw,circle,minimum size=12pt,inner sep=0pt, fill=red!22, draw=red!70] at ($(IS1.east) + (up:30pt) + (right:8pt)$) (P1) {P}
edge[arrows = {-Stealth[scale=2]}, dashed, draw=red] (pd1.north);
\draw[arrows = {-Stealth[scale=2]},draw=blue,dashed] (truck1) ..controls  +(0.0,-2.1).. (etd1);
\draw[arrows = {-Stealth[scale=2]},draw,dashed] (car1) ..controls  +(-1.3,0.0).. (cd1);
\draw[arrows = {-Stealth[scale=2]},draw,dashed] (ego1) ..controls  +(0.0,1.0).. (etd1);

%scenario 3
\node[street2=gray, anchor=west, align=center] at  ($(IS1.east) + (down:40pt) + (right:125pt)$) (IS3) {};
\node[anchor=west, text width=95pt, align=center] at ($(ego1text.north east) + (up:115pt) + (left:20pt)$) (ego2text){\Large (b)};
\node[street=gray, anchor=west, align=center, rotate=90] at  ($(IS3.center) + (up:20pt) + (left:0pt)$) (IS3b) {};

\node[circle,minimum size=12pt,inner sep=0pt] at ($(IS3b.south) + (down:10pt) + (left:20pt)$) (ecd3) {};

\node[ego, anchor=south, align=center] at  ($(IS3.south) + (up:30pt) + (right:10pt)$) (ego3) {};
\node[triangle=black, anchor=north, align=center] at  ($(ego3.north) + (up:0pt) + (left:0pt)$) (ego3b) {};
\node[anchor=north, text width=95pt, align=center] at ($(ego3b.south) + (down:0pt) + (right:0pt)$) (ego3text){E};
\node[ego, anchor=south, align=center, rotate=90] at  ($(IS3.east) + (up:30pt) + (left:60pt)$) (car3) {};
\node[triangle=black, anchor=north, align=center, rotate=-90] at  ($(car3.north) + (up:0pt) + (right:26pt)$) (car3b) {};
\node[anchor=north, text width=95pt, align=center] at ($(car3b.east) + (up:12pt) + (left:10pt)$) (car3text){C};

\draw[arrows = {-Stealth[scale=2]},draw,dashed] (ego3) ..controls  +(0.0,1.0).. (ecd3);
\draw[arrows = {-Stealth[scale=2]},draw,dashed] (car3) -- (ecd3);

\end{tikzpicture}
	\end{adjustbox}
    \caption{Two experimental scenarios: a) ego turns right, b) ego turns left. C: car, E: ego vehicle, P: pedestrian, T: truck.}
    \label{fig:scenarios}
\end{figure}
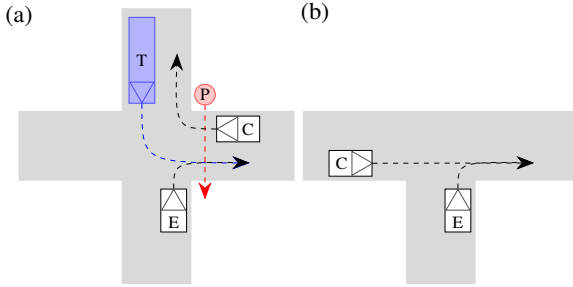
We ran two different experimental setups, each with another logical scenario.

\subsubsection{Scenario 1}
In \Cref{fig:scenarios}, the first logical intersection scenario is presented, which consists of multiple actors, including an ego vehicle (E), a pedestrian (P), a second car (C), and a truck (T) located at an X-intersection. 
The ego vehicle makes a right turn, crossing the trajectories of the pedestrian and the truck but not the car C's trajectory.
During the course of the scenario, both the truck and the second car obstruct the ego vehicle's view of the pedestrian. 
The parameters that were varied in this scenario include the starting s-coordinate of the ego vehicle, the duration of the pedestrian is waiting before crossing the street, and the speed of the car C.

In this scenario, the ego vehicle is the only actor that does not simply follow a predefined trajectory, while all other actors do not possess their own behavioral models.
As simulator, IPG CarMaker was used.

\subsubsection{Scenario 2}
The second scenario occurs at a T-intersection where the ego vehicle (E) makes a right turn while a second car (C) moves straight across the intersection.
Both vehicles have synchronization points, and the second car attempts to reach its synchronization point at the same time as the ego vehicle.
The scenario includes varied parameters such as the ego vehicle's synchronization point coordinates, the starting time for the second car C's movement, and the distance the ego vehicle must travel before the second car's routing is initialized.
The car C has its own (simple) behavior model.
In the course of the experiments, the behavior of the ego vehicle was subject to variation, using three different behavior models from the Carla simulator \cite{Dosovitskiy17}, i.e., cautious, normal, and aggressive, across three different trials conducted with this logical scenario.
All simulations of scenario 2 were executed in the Carla Simulator.

The two scenarios under consideration exhibit distinct characteristics regarding their parameters and physics simulation. 
In the first scenario, the parameters have a transparent and interpretable relationship with the actors' behaviors and actions, as they involve attributes such as speed and position that are easily understood by humans.
Conversely, the second scenario introduces the possibility of hidden parameters that are influenced by the behavior of the ego vehicle and the interactions among actors, adding a layer of complexity to the scenario.

Another notable difference pertains to the physics simulation.
CarMaker, the simulation platform used in the first scenario, does not incorporate a comprehensive physics engine, allowing actors to pass through each other without collision simulation.
Carla is implemented in the game engine Unreal 4, and uses its physics engine to simulate crashes.
Consequently, in scenario 2, all simulations cease as soon as a crash is recorded, whereas in CarMaker, actors successfully reach their designated destination points after each simulation run, and collisions can only be recognized by the utilized criticality metrics.
%The cluster containing scenarios where the ego vehicle crosses before the pedestrian could be reduced even further since in this cluster only the ones close to the criticality border yield yield interesting information.

\subsection{Optimization Problem}

In the first scenario, the criticality between the ego vehicle and the pedestrian, considered the most vulnerable road user in this context, is optimized.
In the second scenario, the focus is on optimizing the criticality between the ego vehicle and car C.
The criticality between actors is assessed using criticality metrics, and the optimization process is carried out to determine the extent to which logical scenarios contain potentially critical situations. 
For scenario 1, two metrics were used as objective functions for Bayesian optimization:
\begin{itemize}
    \item \textbf{Trajectory Distance}: Distance between two actors along their trajectories and road network.
    \item \textbf{Worst-time-to-collision} \cite{wachenfeld_worst-time--collision_2016}: Metric based on TTC \cite{hayward_near_1972}, but without the TTC's limitation to car following scenarios. It assumes the worst case for two actors in terms of a critical encounter.
\end{itemize}
For scenario 2:
\begin{itemize}
    \item \textbf{Euclidean Distance}: Direct distance in space between two actors. 
    \item \textbf{Traffic Quality}: The overall assessment of the traffic within the scene the ego vehicle is situated \cite{schutt2023inverse}.
    \item \textbf{Worst-time-to-collision}.
\end{itemize}
With the exception of traffic quality, all metrics listed above require to find the smallest value to optimize criticality within logical scenarios. 
In contrast, maximizing traffic quality is necessary to achieve this goal.

\subsection{Scenario Clustering}
%After simulation, the gained data needs to be process for further computations.
\subsubsection{Scenario Exploration}
In the first step, the original parameter space has to be reduced from a potentially infinite number of scenarios to a group of representing scenarios.
The simplest solution would be a grid over the parameter space, however, a grid with a sufficient small step size still grows exponentially with new parameters.
To overcome this problem we suggest Bayes Optimization with Gaussian Processes to explore the parameter space \cite{schutt2022application}.
This procedure is a sequential model-based approach that leverages probabilistic surrogate models to efficiently search for the optimal solution in a black-box optimization problem by iteratively selecting promising points based on the acquisition function.

\subsubsection{Computation of Distance}
\label{sec:distance_matrix}
As the spatial distribution of scenarios in the parameter space does not provide any meaningful insights into the scenario outcome, and the selection of parameter combinations is determined by the optimization algorithm which draws from parameter ranges regarding its exploration and exploitation model, it is necessary to establish scenario similarity using other measures than distance in the parameter space.
To compute the similarity of scenarios, dynamic time warping (DTW) \cite{bellman1959adaptive} was utilized as a distance measure for vehicle trajectories or course of the criticality metrics during a scenario.
DTW was chosen since it can handle temporal distortions, variation in the duration and the speed of trajectories.
In many simulates scenarios, the ego vehicle's trajectory remains mostly the same. 
However, it has to react (and brake) at different times during the simulation due to different adversary actor behavior.

\subsubsection{Dimensionality Reduction}
The distance matrix was utilized further to compute a Gaussian radial basis function (RBF) kernel for a kernel principal component analysis (PCA).
Kernel PCA projects the original data into a higher-dimensional feature space so that a linear PCA can be applied \cite{mika1998kernel}.
Therefore, this technique can achieve a dimensionality reduction for a non-linear feature space. 
Two different kernels are suggested by Mika \textit{et al.} \cite{mika1998kernel}: Gaussian and polynomial. 
Due to insufficient data and knowledge about our scenario space as discussed in \Cref{sec:distance_matrix}, only the Gaussian RBF kernel can be computed.
In general, Gaussian RBF kernels are a popular choice since they only have one parameter (kernel width) for hyperparameter tuning and are flexible enough to capture non-linear relationships between data points.

\subsubsection{Clustering}
Density Based Spatial Clustering of Applications with Noise (DBSCAN) \cite{ester1996density} was used for clustering scenarios as an unsupervised clustering method that does not require prior knowledge of the number of clusters.
An advantage of DBSCAN is that it can find clusters of any shape, and unlike k-means, it does not assume clusters to be convexly shaped.
It only has two input parameters that require hyperparameter tuning, the minimum number of points for one cluster and the size of the neighborhood of one cluster.

\subsubsection{Archetype Analysis}
Archetypes are unique data points or extreme examples within a cluster that represent distinct patterns or characteristics. 
%They can be visualized as the \textit{corners} or outer points of a cluster. 
In theory, the proximity of a data point to the center of the cluster indicates its degree of similarity to the archetypes. 
Data points closer to the center are more of a combination or mixture of the archetypal patterns.
For convex clusters, it is possible to identify these archetypes via a principal hull analysis \cite{cutler1994archetypal}.
The principal hull represents the outer boundary of the cluster, and the archetypes can be found at the vertices of this hull.
In the case of a logical scenario, it is assumed that these archetypes are scenarios with the highest variability in actors' behavior and criticality.

\subsection{Discussion}
%\begin{figure*}[tbh]
%    \centering
%       \includegraphics[width=0.92\linewidth]{img/Eval_1.pdf}
%\caption{tbd}
%\label{fig:eval_scenario1}
%\vspace{0ex}
%\end{figure*}

\subsubsection{Scenario 1}

%\begin{figure*}[tbh!]
%    \centering
%     \def\svgwidth{0.99\linewidth}
%     \input{Eval_1_pdf.pdf_tex}
%\caption{(a)-(c) Projections into different 3D kernel spaces.
%(d) Principal Hull Analysis (PHA) to identify archetypal scenarios.
%(e)-(g) Clustering from kernel spaces visualized in scenario parameter spaces.
%(h) Criticality with respect to Worst-Time-to-Collision (WTTC) for each scenario.}
%\label{fig:eval_scenario1}
%\vspace{-3ex}
%\end{figure*}
\begin{figure*}[tbh!]
    \centering
     \includegraphics[width=0.99\linewidth]{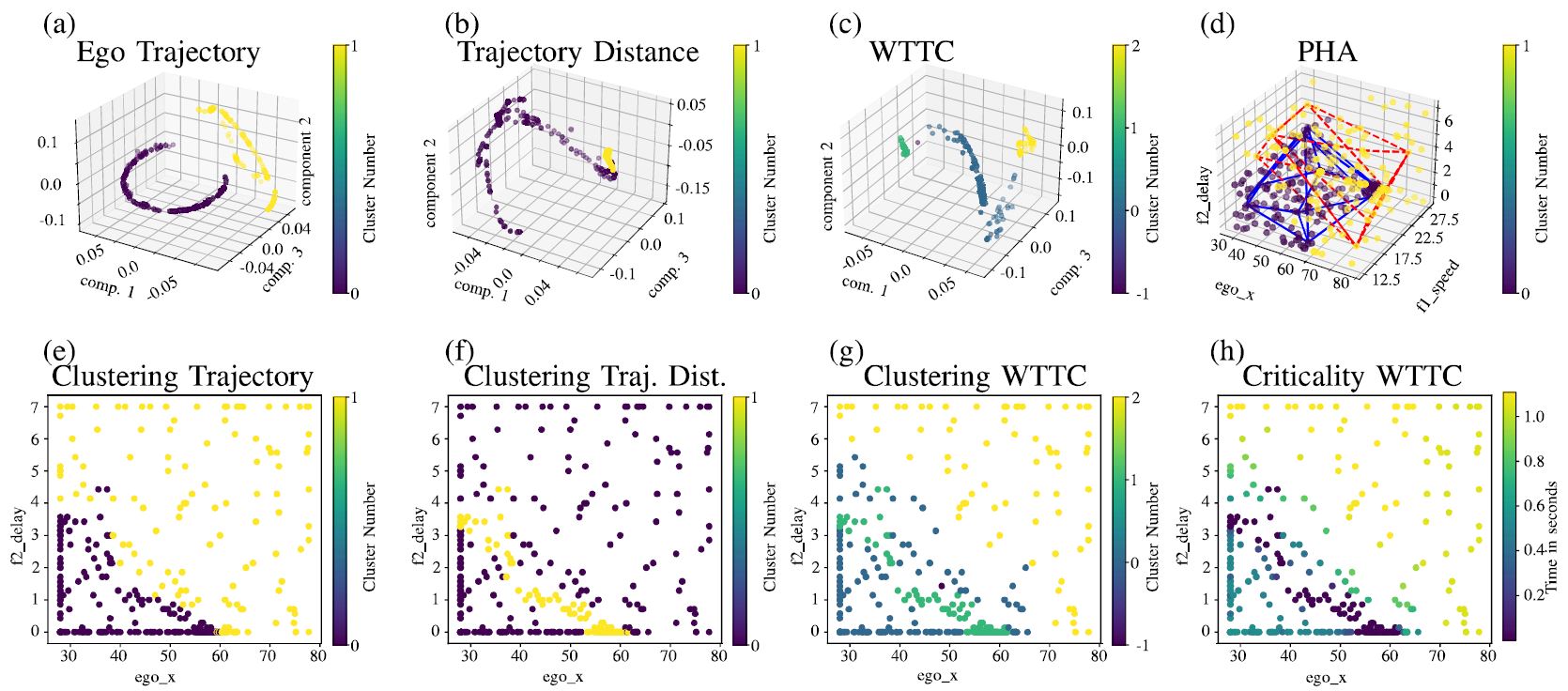}
\caption{(a)-(c) Projections into different 3D kernel spaces.
(d) Principal Hull Analysis (PHA) to identify archetypal scenarios.
(e)-(g) Clustering from kernel spaces visualized in scenario parameter spaces.
(h) Criticality with respect to Worst-Time-to-Collision (WTTC) for each scenario.}
\label{fig:eval_scenario1}
%\vspace{-3ex}
\end{figure*}

The exploration of scenario 1 involved approximately 400 executions of concrete scenarios within the logical scenario.
Remarkably, one of the three given parameters, $p1\_speed$, had negligible influence on the scenario outcome. 
Furthermore, all criticality metrics exhibited similar patterns within the scenario space, with each metric identifying the same critical diagonal in the bottom left corner.
For a more detailed discussion about the results regarding different criticality metrics and the exploration process, we refer readers to \cite{schutt2022application}.
Other than in the previous paper, we primarily focus on scenario clustering to reduce the number of concrete scenarios for testing purposes.
%Upon examining \Cref{fig:eval_scenario1} (h), it is important to observe the distinct characteristics of scenarios located below, around, and above the critical diagonal in the bottom left corner.

\Cref{fig:eval_scenario1} (a)-(c) display the projected concrete scenarios in the kernel space, respectively.
Additionally, \Cref{fig:eval_scenario1}(d)-(h) illustrate all simulated concrete scenarios in the original parameter space, with (d) presented in a three-dimensional space and (e)-(h) displayed in a two-dimensional space to better visualize the scenarios, given the negligible influence of $p1\_speed$.
\Cref{fig:eval_scenario1} (h) shows the worst-time-to-collision (WTTC) criticality of all simulated concrete scenarios within the scenario parameter space.
Scenarios below the critical diagonal indicate instances where the ego vehicle passes the intersection after the pedestrian. 
Near the critical diagonal, scenarios depict situations where the car either collides with or closely approaches the pedestrian. 
Finally, above the diagonal are scenarios in which the pedestrian crosses the intersection after the ego vehicle.
Among these three categories, scenarios falling above the diagonal are generally considered less significant as they involve minimal interaction between the ego vehicle and pedestrian, thereby suggesting a potential for reducing the number of concrete scenarios.

\Cref{fig:eval_scenario1} (a) depicts the results of the scenarios clustered regarding the ego vehicle's trajectory (behavior-based) in the kernel space after performing kernel PCA ($\gamma=100.0$). 
At the same time (e) shows the found clusters in the original space.
Interestingly, the cluster borders converge at the area with the highest criticality.
Pairs (b) and (f) represent the kernel space where the trajectory distance was utilized as a distance metric for computing the kernel matrix, while (c) and (g) represent the kernel space using WTTC as the distance metric.
Clustering based on trajectory distance identifies two distinct clusters: high-criticality scenarios and low-criticality clusters.
On the other hand, WTTC can differentiate between high criticality, medium criticality, and low criticality.
A comparison of (g) and (h) demonstrates that the clustering closely aligns with the measured criticality gradient throughout the simulation.
\Cref{fig:eval_scenario1} (d) shows an additional principal hull analysis (red and blue vertices and eges) that encompasses both scenario clusters in the scenario parameter space.
The resulting archetypes of the hull analysis are a suitable foundation for a reduced scenario set, which can be supplemented with prototypical scenarios from the center of each cluster.
For scenario 1 with 15 scenarios defining the hull of each cluster, this leads to 32 scenarios as a reduces set representing the logical scenario 1.

\subsubsection{Scenario 2}
%\begin{figure*}[tbh]
%    \centering
%       \includegraphics[width=0.92\linewidth]{img/platzhalter.png}
%\caption{Platzhalter!!!}
%\label{fig:eval_scenario2}
%\vspace{0ex}
%\end{figure*}

%\begin{figure*}[tbh!]
%    \centering
%     \def\svgwidth{0.99\linewidth}
%     \input{Eval_2.pdf_tex}
%\caption{(a)-(c) Concrete scenarios on the original scenario space, colored by their assigned clusters.
%(d)-(f) Projections into the 3D kernel spaces, depending on the chosen distance matrix, and colored by cluster assignment.}
%\label{fig:eval_scenario2}
%\vspace{-3ex}
%\end{figure*}
\begin{figure*}[tbh!]
    \centering
     \includegraphics[width=0.99\linewidth]{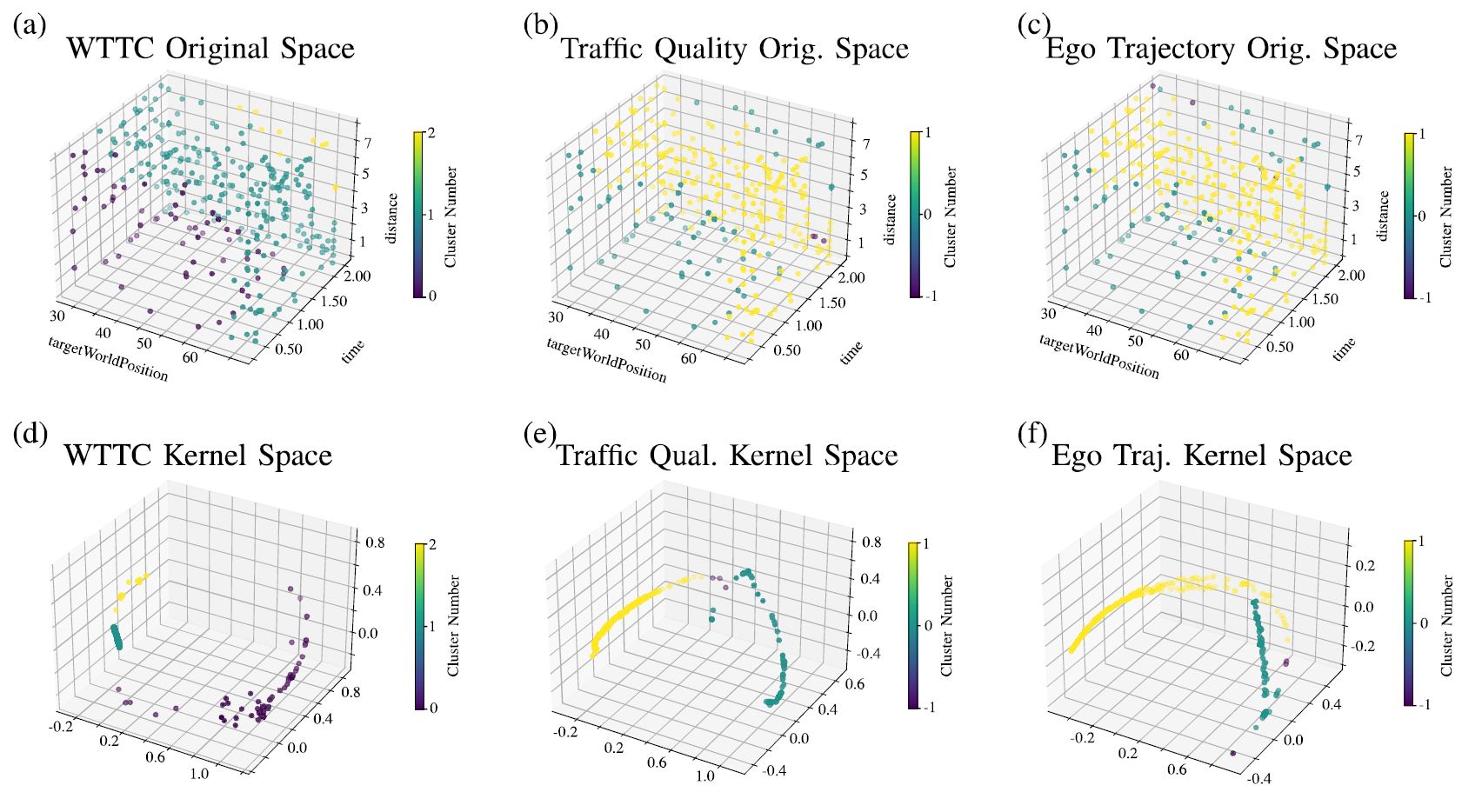}
\caption{(a)-(c) Concrete scenarios on the original scenario space, colored by their assigned clusters.
(d)-(f) Projections into the 3D kernel spaces, depending on the chosen distance matrix, and colored by cluster assignment.}
\label{fig:eval_scenario2}
%\vspace{-3ex}
\end{figure*}

The exploration of Scenario 2 entailed the execution of approximately 300 concrete scenarios. 
In contrast to Scenario 1, all three parameters exhibited an influence on the scenario outcomes, as depicted in \Cref{fig:eval_scenario2}. 
Scenario 2 demonstrated three distinct outcomes: (1) Vehicle C crossing the intersection before the ego vehicle, corresponding to scenarios classified as less critical by the applied criticality metrics with parameters $time$ and $distance$  close to zero, (2) collisions between the two vehicles, and (3) the ego vehicle passing before Vehicle C, represented by scenarios located in the top right corner of the parameter space with high $time$ and $distance$  values.

In \Cref{fig:eval_scenario2}, all plots (a)-(f) of Scenario 2 show a critical region marked as cluster 1. 
This critical region encompasses scenarios with both ego vehicle-caused accidents and accidents caused by vehicle C. 
Notably, critical scenarios in proximity to the region where vehicle C crosses first involve the ego vehicle colliding with vehicle C, which then transitions into accidents caused by vehicle C. 
Unlike Scenario 1, the clustering process, employing dynamic time warping (DTW) to measure distances between ego trajectories, distinguishes accidents as a separate cluster due to the simulation terminating upon a crash. Consequently, three main clusters emerge from the clustering results: (1) non-crash situations represented by clusters 0 and 2 (if existing), (2) accidents at the intersection denoted by cluster 1.
In \Cref{fig:eval_scenario2} (d), the accident free scenarios are divided into two separate clusters, unlike (e) and (f).

The critical region, identified as cluster 1 in the parameter space in (a)-(c) (blue in (a) and yellow in (b) and (c)), encompasses scenarios where accidents occur.
Particularly, scenarios in the vicinity of the region where Vehicle C crosses first involve the ego vehicle colliding with vehicle C, leading to subsequent accidents caused by vehicle C. 
The clustering process using DTW distances between ego trajectories effectively discerns accidents as a distinct cluster, distinguishing them from non-crash scenarios, which differs from the clustering outcome observed in Scenario 1. 
In summary, the clustering analysis yields two primary clusters: (1) non-crash situations and (2) accidents at the intersection

Further insights can be gained from the kernel space. 
In \Cref{fig:eval_scenario2}, all critical scenarios are closely situated together. 
Moreover, in the kernel space where traffic quality was employed, clusters 0 and 1 exhibit the most distinct boundaries. 
This observation is attributed to traffic quality exhibiting a steep gradient in areas where critical and non-critical clusters intersect, thereby enabling clear demarcation between the clusters.

A principal hull analysis was not conducted on the clusters of Scenario 2 due to the non-convex nature of the identified critical clusters.
This opens up possibilities for future research aimed at developing methods to reduce the scenario set for such non-convex cluster shapes.

\subsubsection{Comparison to other methods}
One advantage of the chosen Bayes optimization is the fact that, unlike other algorithms, e.g., genetic algorithms, it holds information about the confidence of the criticality at the areas around simulated data points (concrete scenarios).
However, even though we have this information, it is based on assumptions that are made at the model design, e.g. choosing a Gaussian RBF kernel.
Another possible approach is using a grid for parameter combinations.
However, this approach has a exponentially growing calculation time when additional parameters are introduced into the scenario space.
Also, as Mori \textit{et al.} \cite{mori2022inadequacy} state, there exists a possibility to overlook critical scenarios.
Additionally, in a scenario space defined by three parameter ranges, where 15 steps per parameter are taken, the simulation effort includes 3375 scenarios, more than ten times as many as the optimization needed for scenario 2.
This also leads to growing expanses for the evaluation regarding the clustering.
For approximately 300 concrete scenarios it takes about 3 hours to calculate the distance matrix for the criticality metrics, and it has an exponentially growing computation time: for 3375 it takes 15 days.
For the distance matrices of the criticality metrics it even takes around 363 days.

\section{Conclusion}
\label{sec:conclusion}
In this work, we utilized Bayes optimization with Gaussian processes to explore the scenario space of a logical scenario.
As the basis for the acquisition function, different criticality metrics, i.e., distance, WTTC, and traffic quality, were used.
To gain deeper insights and understanding the characteristics of the scenario space, we adopted a behavior-based and a criticality-based approach for scenario clustering.
These approaches provided different perspectives on the scenario outcomes and offered valuable information regarding the behavior and criticality of different scenarios. 
The resulting scenario clusters served as a foundation for scenario reduction, utilizing archetype analysis.
We were able to identify and extract key representative scenarios, significantly reducing the overall set of distinct scenarios. 
The reduction process successfully minimized the originally simulated data to one-tenth of its original size, enabling a more manageable and focused set of scenarios for further analysis and testing.

In the future, the found scenario set has to be evaluated regarding its coverage of the logical scenario and its probability to overlook unknown critical scenarios.
Additionally, a way to identify archetypes of non-convex clusters has to be investigated.

%%%%%%%%%%%%%%%%%%%%%%%%%%%%%%%%%%%%%%%%%%%%%%%%%%%%%%%%%%%%%%%%%%%%%%%%%%%%%%%%
%\section*{APPENDIX}


\section*{ACKNOWLEDGMENT}
The research leading to these results is funded by the German Federal Ministry for Economic Affairs and Climate Action within the projects \textit{Verifikations- und Validierungsmethoden automatisierter Fahrzeuge im urbanen Umfeld} a project from the PEGASUS family\footnote{\url{https://pegasus-family.de/}}, based on a decision by the Parliament of the Federal Republic of Germany. The authors would like to thank the consortium for the successful cooperation (19A19002L).
%The preferred spelling of the word ÒacknowledgmentÓ in America is without an ÒeÓ after the ÒgÓ. Avoid the stilted expression, ÒOne of us (R. B. G.) thanks . . .Ó  Instead, try ÒR. B. G. thanksÓ. Put sponsor acknowledgments in the unnumbered footnote on the first page.

\section*{Copyright}
\textcopyright 2023 IEEE.  Personal use of this material is permitted.  Permission from IEEE must be obtained for all other uses, in any current or future media, including reprinting/republishing this material for advertising or promotional purposes, creating new collective works, for resale or redistribution to servers or lists, or reuse of any copyrighted component of this work in other works.

%%%%%%%%%%%%%%%%%%%%%%%%%%%%%%%%%%%%%%%%%%%%%%%%%%%%%%%%%%%%%%%%%%%%%%%%%%%%%%%%

\bibliographystyle{IEEEtran}
\bibliography{root}

\end{document}